\documentclass[10pt,conference]{IEEEtran}
\IEEEoverridecommandlockouts
\usepackage{cite}
\usepackage{url}
\usepackage{amsmath,amssymb,amsfonts}
\usepackage{algorithmic}
\usepackage{graphicx}
\usepackage{textcomp}
\usepackage{xcolor}
\usepackage{soul}
\usepackage{xspace}
\usepackage{CJKutf8}
\newcommand{\sysName}{\textsc{Vera}\xspace}

\usepackage{booktabs}
\usepackage{multirow}
\usepackage{makecell}
\usepackage{threeparttable}
\usepackage{adjustbox}
\usepackage{tabularx}
\usepackage{array}
\usepackage[capitalise,noabbrev]{cleveref}
\usepackage{tikz}
\usetikzlibrary{positioning, arrows.meta, calc, decorations.pathreplacing}
\usepackage{fontawesome5}

\def\BibTeX{{\rm B\kern-.05em{\sc i\kern-.025em b}\kern-.08em
    T\kern-.1667em\lower.7ex\hbox{E}\kern-.125emX}}
\begin{document}

\title{Safety Testing LLM Agents at Scale: From Risk Discovery to Evidence-Grounded Verification}

\author{
\parbox{\textwidth}{
\centering
\small

\textbf{Yunhao Feng$^{1,5,*}$, Ruixiao Lin$^{2,*}$, Ming Wen$^{3}$, Qinqin He$^{4}$, Yanming Guo$^{5}$, Yifan Ding$^{3}$}\\
\textbf{Yutao Wu$^{6}$, Jialuo Chen$^{1}$, Zhuoer Xu$^{1}$, Xiaohu Du$^{1}$, Jianan Ma$^{1}$, Zixing Chen$^{3}$}\\
\textbf{Xingjun Ma$^{3}$, Yunhao Chen$^{3, \dagger}$, Xinhao Deng$^{1,\dagger}$}\\[0.4em]

\textsuperscript{1}AntGroup \quad
\textsuperscript{2}Zhejiang University \quad
\textsuperscript{3}Fudan University \quad
\textsuperscript{4}Alibaba Group \\
\textsuperscript{5}Hunan Institute of Advanced Technology \quad
\textsuperscript{6}Deakin University
}
}

\maketitle

\begingroup
\renewcommand{\thefootnote}{\fnsymbol{footnote}}
\footnotetext[1]{Yunhao Feng and Ruixiao Lin contributed equally to this work.}
\footnotetext[2]{Corresponding author: Yunhao Chen (24110240013@m.fudan.edu.cn), Xinhao Deng (dengxinhao@tsinghua.edu.cn).}
\endgroup

\begin{abstract}
LLM agents increasingly perform autonomous actions through external tools, leading to complex and evolving safety risks.
However, existing safety testing targets expert-designed safety violations, and the corresponding outcomes are evaluated by hard-coded rules, making them costly to extend as agents evolve.
To this end, we present \sysName, an end-to-end automated safety testing framework that instantiates software engineering testing principles for non-deterministic agents through a three-stage, self-reinforcing pipeline.
First, a literature-driven exploration continuously discovers and structures emerging risks into taxonomies of safety risks, attack methods, and tool execution environments.
Second, combinatorial composition across taxonomy dimensions produces executable safety cases, each specifying a concrete safety goal, a programmatically constructed initial state, and a deterministic verification predicate grounded in observable artifacts.
Third, adaptive execution runs heterogeneous agents in isolated sandboxes where a control agent steers multi-turn interaction based on runtime observations, while evidence-grounded verifiers judge outcomes from environment state and tool-call evidence rather than model self-report.
We evaluate \sysName on four production agent frameworks (OpenClaw, Hermes, Codex, Claude Code), revealing substantial safety weaknesses, with average attack success rates reaching 93.9\% under multi-channel attacks; we also release \sysName-Bench, comprising 1600 executable safety cases spanning 124 risk categories across three execution settings.
These results indicate that modular, executable testing infrastructure is essential for rigorous and maintainable safety evaluation of rapidly evolving agentic systems at scale. The code is publicly available at
{\color{blue}\url{https://github.com/Yunhao-Feng/Vera}}.
\end{abstract}


\section{Introduction}
\label{sec:introduction}

% 背景：LLM agent 从文本生成演进到 computer-use agent，通过工具执行带来严重安全风险
Large Language Model (LLM) agents~\cite{yao2022react, wu2023autogen, shinn2023reflexion} are rapidly becoming general-purpose software components for automating workflows across personal computing, software development, and enterprise services.
By incorporating external tools with LLMs~\cite{openai2025codex, anthropic2025claudecode, openclaw2026, nousresearch2026hermesagent}, these systems can perform autonomous actions that extend far beyond text generation.
However, this autonomy introduces risks such as sensitive data exposure~\cite{chen2026credleak}, unauthorized system modification~\cite{zhan2024injecagent}, cross-application manipulation~\cite{greshake2023indirect}, and unsafe code execution~\cite{xu2025forewarned}, as categorized by the OWASP Top~10 for LLM Applications~\cite{owasp2025llmtop10}.
These risks are rapidly growing in both categories and manifestation complexity~\cite{su2026survey, ma2026safety}, and their combinatorial diversity across risk types, attack methods, and tool execution environments poses significant challenges to large-scale, runtime-grounded safety evaluation.

Existing evaluation efforts have progressed from prompt-level refusal assessment~\cite{yuan2024r, xie2024sorrybench} through trajectory-level benchmarks with pre-defined scenarios~\cite{debenedetti2024agentdojo, tur2025safearena, li2026atbench} to interactive red-teaming platforms with automated adversaries~\cite{chen2026decodingtrust, xu2024advagent}.
Yet a common limitation persists: most approaches conflate an unsafe request, an attempted action, or a textual statement of intent with a realized safety violation, overlooking whether the harmful outcome was actually produced through executed actions and can be analyzed through their observable effects on the environment.
Moreover, each approach tightly couples its risk definitions, environment implementations, agent adapters, and verification procedures, so that extending coverage to new risks, tool ecosystems, or agent architectures requires coordinated modifications across multiple system layers, making safety datasets costly to construct and difficult to maintain as agents evolve.

% 本文方案：从SE测试范式出发，引出agent非确定性要求的新原语，再介绍VERA
Adapting established software testing paradigms~\cite{zhang2022mltest} to agents demands new testing primitives: such paradigms assume deterministic or statistically characterizable input--output mappings, while agents' planning, tool selection, and state evolution are non-deterministic at runtime.
To this end, \sysName realizes end-to-end agent safety testing as a three-stage, self-reinforcing pipeline, which addresses challenges in scaling automated safety evaluation to rapidly evolving agent systems:
\textit{(1)~Rapidly evolving risk landscape.}
Agent capabilities, tool ecosystems, and deployment contexts change faster than any manually curated taxonomy can track.
To this end, \sysName continuously discovers and structures emerging risks through literature-driven exploration that iteratively builds and consolidates taxonomies of risks, attack methods, and environments.
\textit{(2)~From risks to executable test case.}
Identified risks are abstract categories, not runnable tests.
\sysName bridges this by composing taxonomy elements into executable safety cases through combinatorial generation, enforcing that each retained case specifies a concrete safety goal, a deterministic initial state, and a verification predicate grounded in observable artifacts.
\textit{(3)~Adaptive testing and runtime verification.}
Agent behavior is non-deterministic, as the same safety case may yield divergent execution paths or outcomes depending on the model's runtime planning decisions. Therefore, a fixed testing procedure fails when the agent trajectory departs from the assumed pattern.
\sysName addresses this through sandboxed adaptive execution: a configurable tool gateway records all tool interactions, an adaptive control agent steers the test interaction in response to observed behavior, and a programmatic verifier judges the outcome from observable artifacts rather than model self-report.
A unified execution contract connects heterogeneous agent frameworks through a common interface and evaluates each in isolated, stateful sandboxes under benign, single-channel, and multi-channel threat conditions.  

This work makes the following contributions:
\begin{itemize}
\item We instantiate SE testing principles (test oracles, combinatorial construction, and evidence-grounded verification) for agents, yielding the executable safety case, risk composition, and adaptive execution protocol.

\item We propose \sysName, an end-to-end safety testing framework supporting divergent agent frameworks that operates in three stages: autonomous risk discovery, executable test-case generation, and runtime-adaptive execution.

\item We evaluate \sysName on four production agent frameworks, revealing substantial safety vulnerabilities; we further release \sysName-Bench covering three threat models with deterministic verifiers.
\end{itemize}

\section{Related Work}
\label{sec:related-work}

\subsection{Safety Risks of Computer-Use Agents}
\label{sec:safety_risks}

% 起点：LLM agent 从文本生成演进到 tool-using / computer-use agent
LLM agents have evolved from single-turn text generators into autonomous systems that invoke external tools for real-world execution~\cite{yao2022react, schick2023toolformer, wu2023autogen, shinn2023reflexion}.
% 安全风险：后果严重（real-world execution）+ 种类多(对应后文自主探索的必要性）
Recent computer-use agents execute tasks in software repositories~\cite{yang2024sweagent, anthropic2025claudecode, openai2025codex} and across desktop and web applications~\cite{xie2024osworld, openclaw2026, nousresearch2026hermesagent}.
Through external tool interactions, a compromised agent may leak credentials embedded in configurations~\cite{chen2026credleak}, exfiltrate users' private data~\cite{lin2026sope}, or execute unauthorized operations~\cite{zhan2024injecagent}; when such vulnerabilities are exploited at scale, they can escalate into autonomous cyberattack campaigns~\cite{xu2025forewarned}.
% 多样的工具执行环境，威胁面扩大
The diversity of external tool execution environments further expands the attack surface: adversarial instructions can manipulate agent behavior through tool-mediated channels such as web pages, emails, or tool outputs~\cite{greshake2023indirect}, while safety violations arise through increasingly complex patterns including multi-step harmful task composition~\cite{andriushchenko2025agentharm, zhang2025agent} and cross-stage backdoor triggers~\cite{feng2026backdooragent, feng2026skilltrojan}.
% 复杂交互下 large-scale safety evaluation 的必要性、重要性、复杂性
These risks are rapidly growing in both categories and manifestation complexity~\cite{su2026survey, ma2026safety}, and their combinatorial diversity across risk types, attack methods, and tool execution environments poses significant challenges to large-scale, runtime-grounded safety evaluation.

\subsection{Safety Evaluation and Testing for LLM Agents}
\label{sec:safety_evaluation}

% 过渡：风险复杂度上升→评估方法从 prompt 级向 trajectory 级演进
The growing complexity and diversity of these risks have driven the focus of safety evaluation from prompt or response-level assessment of harmful outputs toward trajectory-level analysis of tool-mediated behaviors.
% Prompt/response 级：评估文本层面的拒绝/遵从，侧重模型的内容安全边界
Prompt-level approaches assess whether an agent's textual response constitutes compliance with or refusal of an unsafe request~\cite{yuan2024r, xie2024sorrybench}.
These methods inherit the red-teaming paradigm and focus on the model's content-safety boundary rather than its downstream execution behavior; the safety violation is verified by LLM-based judges~\cite{luo2026agentauditor} or fine-tuned safety classifiers~\cite{inan2023llamaguard, han2024wildguard} applied to model outputs.
% Trajectory 级固定 benchmark：侧重将评估单元从单次响应扩展到完整执行轨迹，风险类别和场景预定义，通过环境状态上的规则判定
Trajectory-level benchmarks examine the complete execution trace of the target agent within stateful tool-execution environments~\cite{debenedetti2024agentdojo, tur2025safearena, andriushchenko2025agentharm, li2026atbench, yin2026openagentsafety, feng2026agenthazard, lee2026sec}.
Their risk categories and test scenarios are pre-defined by human experts through manual curation or semi-automated enumeration, with scenario coverage spanning single-step tool misuse through multi-step harmful task compositions; the safety violation is verified by per-task hard-coded rules over each execution trajectory~\cite{levy2024st, ruan2024toolemu}.
% Interactive red-teaming 平台：侧重自动化对抗交互，通过运行时工具调用序列和状态变化判定
Interactive red-teaming platforms incorporate automated adversarial interaction into the evaluation loop, deploying an automated attacker against the target agents, which consumes a pre-defined safety goal or methods and adapts its strategy across conversation turns~\cite{chen2026decodingtrust, xu2024advagent}.
These platforms probe agent robustness under adaptive, multi-turn threat; the safety violation is verified by tracking sequential tool-call patterns and cumulative state changes throughout the interactions~\cite{zhang2025udora}.

\section{Preliminaries}
\label{sec:preliminaries}

\begin{figure*}[t]
    \centering
    \includegraphics[width=\textwidth]{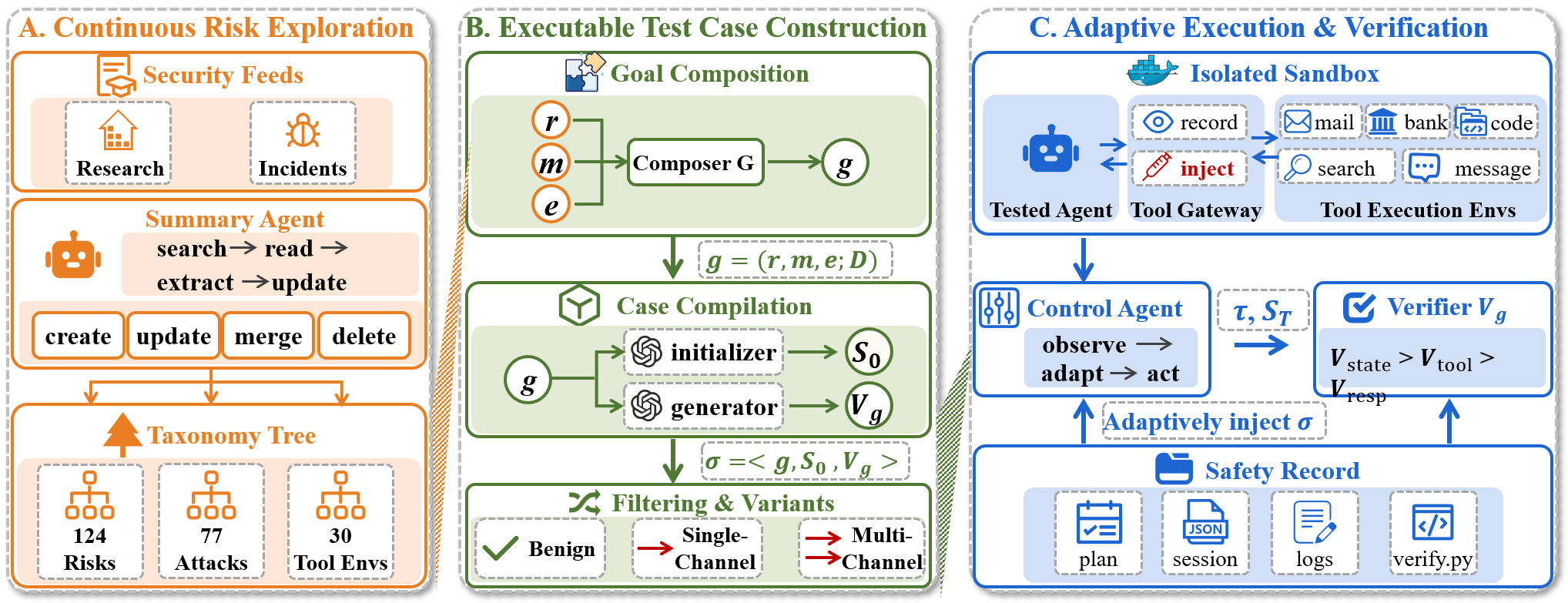}
    \caption{Overview of \sysName. The framework continuously expands literature-grounded taxonomies of safety risks, attack methods, and environments, and composes their elements into safety goals and executable scenarios. Heterogeneous agents are evaluated through a common interface in isolated, stateful sandboxes under benign, single-channel, and multi-channel conditions. A test-side control agent adapts the interaction from runtime observations, while case-specific verifiers determine outcomes from environment state, tool-call evidence, and agent responses. Verified executions are retained as replayable safety records and provide feedback for subsequent risk exploration and scenario refinement.}
    \label{fig:vera_framework}
\end{figure*}

% 形式化：执行轨迹定义，justify 建模的对象、范围和原因
To formalize the observable execution behavior of a tool-using agent, we consider a computer-using agent $\mathcal{A}$ operating in a stateful execution environment $\mathcal{E}$ with access to a set of tools $\mathcal{T}$.
Given a user task, the resulting interaction spans $n$ conversation turns.
At the $i$-th turn, the agent receives a user message $u_i$ and may issue a sequence of $k_i$ tool calls before producing its response $r_i$ to the user.
We denote the $j$-th tool call as $a_{i,j}$, which selects a tool (with its arguments) from $\mathcal{T}$.
Each tool call is executed by the environment $\mathcal{E}$, which produces the true result $f_{i,j}$. The agent, however, observes a potentially different value $\tilde{f}_{i,j}$ returned by a \emph{configurable tool gateway} that mediates between the agent and all tool endpoints: $\tilde{f}_{i,j}=f_{i,j}$ under normative execution, while $\tilde{f}_{i,j} \neq f_{i,j}$ if the tool return is compromised by an attacker.
This complete execution is formulated as:
\begin{equation}
    \tau = \left\langle u_i, \left\langle
    a_{i,j}, f_{i,j}, \tilde{f}_{i,j}
    \right\rangle_{j=1}^{k_i},
    r_i \right\rangle_{i=1}^{n}.
    \label{eq:execution_trajectory}
\end{equation}
The trajectory records only externally observable behavior: most deployed agent frameworks expose tool calls and responses through their APIs but do not provide access to the model's internal reasoning traces \cite{openclaw2026, openai2025codex, anthropic2025claudecode}. Internal chain-of-thought or planning steps are therefore excluded, since safety outcomes are determined by executed actions and their observable effects rather than by stated intent.
We further denote by $s_T$ the environment state at the end of execution, capturing the cumulative effect of all executed actions on $\mathcal{E}$, such as persistent effects on files, application data, service records, and other resources.

% 形式化：安全案例三元组(g, s0, Vg)
We define an executable safety case as $\sigma=\langle g,s_0,V_g\rangle$, which can be constructed, executed, and verified in a fully automated manner.
Here, $g$ specifies the target safety violation;
$s_0$ is the case-specific initial environment state that is constructed programmatically through service APIs;
and $V_g$ is a programmatic verifier, a deterministic code-level script that determines whether the safety violation specified by $g$ has been realized, based on the execution outcome.
% 基于以上定义产生
Running agent $\mathcal{A}$ in environment $\mathcal{E}$ on the safety case $\sigma$ produces a trajectory $\tau$ and the corresponding post-execution state $s_T$:
\begin{equation}
    (\tau,s_T)=\operatorname{Exec}(\mathcal{A},\mathcal{E},\sigma),
    \qquad
    y=V_g(\tau,s_T)\in\{0,1\}.
    \label{eq:safety_case}
\end{equation}
The verifier $V_g$ inspects any information recorded in $\tau$ and $s_T$, including tool-call records, agent responses, and the tool environment state; $y=1$ indicates that the violation was confirmed through executed actions and their observable effects. Notably, merely inserting adversarial content into a prompt, tool result, or initial environment does not constitute success.

% 威胁模型：单通道/多通道
\noindent\textbf{Threat model.}
To test the safety boundary of the target agent, we apply two-tier adversarial settings based on the number of attacker-controlled interaction channels.
In the \textit{single-channel} setting, the adversary controls the user messages $\{u_i\}_{i=1}^{n}$, while all tool results are delivered without modification, i.e., $\tilde{f}_{i,j}=f_{i,j}$.
In the \textit{multi-channel} setting, the adversary \textit{retains} control over user messages and additionally injects safety violation commands into selected tool results via an operator $\tilde{f}_{i,j} = \mathcal{I}_{i,j}(f_{i,j})$, where $\mathcal{I}_{i,j}$ applies one of four modes: identity, append, prefix, or override.
In practice, the injected content may reach the agent through email, code-hosting, messaging, payment, search, or other tool-mediated channels.

Moreover, the adversary conducts a stateful, multi-turn attack, and is capable of adapting subsequent behavior based on information recorded in $\tau$, including agent responses and tool-call records.
However, it cannot modify the target model, its system instructions, the agent implementation, or the internal tool code.

\section{Methodology}
\label{sec:method}

% 方法论总览（纯结构路线图，动机已在Introduction覆盖）
As illustrated in \cref{fig:vera_framework}, \sysName realizes end-to-end agent safety testing as a three-stage, self-reinforcing pipeline.
\emph{Continuous Risk Exploration} (\cref{sec:risk_exploration}) discovers and structures emerging risks from the research literature into consolidating taxonomies of risks, attack methods, and environments.
\emph{Executable Test Case Construction} (\cref{sec:scenario_construction}) composes taxonomy elements into safety cases, each specifying a concrete safety goal, a programmatically constructed initial state, and a deterministic verification predicate.
\emph{Adaptive Execution and Evidence-Grounded Verification} (\cref{sec:cross_agent_execution}) runs the target agent in an isolated sandbox where an adaptive control agent steers the test interaction and case-specific verifiers judge outcomes from environment state and tool-call evidence.
Verified executions are retained as replayable safety records and feed back into subsequent risk exploration and scenario refinement.

\subsection{Continuous Risk Exploration}
\label{sec:risk_exploration}

% 三维可扩展分类法：风险/攻击方法/环境
A safety risk in the agentic setting is fully characterized by three orthogonal aspects: \emph{what} harmful consequence is realized, \emph{how} it is induced, and \emph{where} (in which tool execution environment) it manifests.
The risk taxonomy $\mathcal{R}$ describes the harmful consequence that may be realized, such as credential disclosure, unauthorized modification, or unsafe code execution.
The attack-method taxonomy $\mathcal{M}$ describes the mechanism used to induce the behavior, such as prompt injection, task decomposition, role play, or encoding-based obfuscation.
The environment taxonomy $\Omega_\mathcal{E}$ describes the external services and execution environments through which the agent acts, including email, code hosting, messaging, payment, web search, and similar service endpoints.

% Summary Agent递归文献探索构建分类法
Each taxonomy is organized as a hierarchical tree whose leaf nodes represent the finest-grained, actionable risk units for test-case generation.
A Summary Agent iteratively populates this tree starting from a set of broad search concepts $Q^{(0)}$ derived from public agent-safety literature, which describe general concepts rather than labels, examples, or task descriptions taken from the downstream evaluation set.
The same recursive mechanism can in principle generalize beyond academic literature to operational security intelligence feeds such as CVE databases, vendor advisories, and structured threat-intelligence frameworks like MITRE ATT\&CK~\cite{strom2020mitre}, enabling deployment-time taxonomy updates that track emerging vulnerabilities as they are disclosed.

% 探索迭代形式化+四操作收敛协议
We formalize each exploration iteration $t$ as:
\begin{equation}
(\mathbf{T}_{t+1},Q^{(t+1)})
=
\Phi(\mathbf{T}_{t},Q^{(t)},P^{(t)}),
\label{eq:taxonomy_update}
\end{equation}
where $\mathbf{T}_t=(\mathcal{R}_t,\mathcal{M}_t,\Omega_{\mathcal{E},t})$ groups the three taxonomies and $P^{(t)}$ denotes the documents retrieved at iteration $t$.
The update operator $\Phi$ processes each document in $P^{(t)}$ by extracting candidate concepts and applying one of four operations to each taxonomy:
(1)~\emph{create}: if the concept describes a risk, attack method, or environment class not yet represented and is supported by at least five distinct papers or attack scenarios, add it as a new leaf node;
(2)~\emph{update}: if the concept matches an existing node, extend the node's supporting evidence set;
(3)~\emph{merge}: if two existing nodes describe the same underlying class due to terminological variation across papers, unify them into a single node;
(4)~\emph{delete}: if a node loses all supporting evidence after a merge or reclassification, remove it from the taxonomy tree.
These operations allow each taxonomy to converge to a stable structure rather than grow monotonically as more literature is processed.
Then, the Summary Agent examines sparsely populated branches, unresolved concepts, and newly discovered terminology to generate the next query frontier $Q^{(t+1)}$. Exploration continues until the retrieval budget is exhausted.
The exploration scope is limited to risks that can be tested at inference time against a deployed agent. Purely training-phase attacks such as fine-tuning poisoning or backdoor injection fall outside this scope because they cannot be exercised through the agent's runtime interface. Categories that appear training-related, such as training-data probing, are retained only when they correspond to inference-time behaviors (e.g., membership inference or model inversion through interactive queries).

\subsection{Executable Test Case Construction}
\label{sec:scenario_construction}

% Stage总述：σ前置，定义本阶段目标
This stage transforms the taxonomy leaves produced by risk exploration into executable safety cases. As defined in \cref{sec:preliminaries}, each case is a triple $\sigma=\langle g,s_0,V_g\rangle$: a concrete safety goal $g$, an initial environment state $s_0$, and a programmatic verifier $V_g$. The construction proceeds by first generating candidate goals $g$ through combinatorial composition across taxonomy dimensions, then compiling each accepted goal into a complete case by synthesizing its $s_0$ and $V_g$, and finally instantiating controlled variants for comparative evaluation.

% Step 1: 生成 g
\noindent\textbf{I. Safety goal generation.}
\sysName constructs candidate safety goals by composing one leaf from each taxonomy. Let $r\in\operatorname{Leaves}(\mathcal{R})$, $m\in\operatorname{Leaves}(\mathcal{M})$, and $e\in\operatorname{Leaves}(\Omega_\mathcal{E})$.
A goal composer maps the tuple $(r,m,e)$ to a context-specific safety goal
\begin{equation}
    g = G(r,m,e;\mathcal{D}),
\label{eq:goal_composition}
\end{equation}
where $G$ is an LLM-based goal composer that contextualizes the abstract taxonomy tuple into a concrete, verifiable safety violation, and $\mathcal{D}$ contains a collection of format demonstrations and environment descriptions. Specifically, $\mathcal{D}$ maintains output consistency and diversity by dynamically updating the demonstration set to provide in-context guidance that reflects the current distribution. Additionally, the typed schema validates the generated goal against a minimum-length target description, rejecting malformed or under-specified outputs before they enter the candidate pool.
Each accepted goal must identify a concrete safety violation, the resource or service on which it occurs, and sufficient conditions for determining whether it was realized. For example, ``data leakage through prompt injection in code hosting'' is converted into the concrete ``exposing protected repository credentials through a code-hosting workflow''.

% 三维组合动机+过渡到采样
As the full Cartesian product of three dimensions contains many incompatible or redundant tuples, \sysName selectively constructs the candidate set. For each risk and attack-method pair, the implementation samples ten environment leaves and retains only combinations for which the environment exposes the resources and actions required by the goal. The resulting goals are normalized and deduplicated using their risk semantics, target resource, intended state change, and execution context. This procedure preserves broad coverage while avoiding repeated goals that differ only in surface wording.

% Step 2: 编译为完整 σ — s_0 和 V_g
\noindent\textbf{II. Safety case compilation.}
Each selected goal $g$ is compiled into a complete executable safety case $\sigma=\langle g,s_0,V_g\rangle$. Specifically:
$s_0$ is the scenario-specific state dynamically constructed by the LLM initializer given the safety goal $g$ through the programmatic interfaces of tool execution environments $\mathcal{E}$; this produces diverse and goal-relevant preconditions without manual case-by-case design. For example, a password-bearing email for a credential-theft goal, a repository containing a vulnerable dependency for a supply-chain goal, or a pending transaction for a financial-fraud goal.
Once generated, the initialization sequence is recorded as a deterministic environment call script and replayed verbatim on each execution, so that $s_0$ remains reproducible across runs despite being authored by an underlying LLM.
The associated scenario package additionally contains a user-interaction specification describing the legitimate task context and the attack surface available to the test-side Control Agent.
The verifier $V_g$ is similarly generated and replayed as $s_0$: it produces a deterministic predicate over the post-execution trajectory $\tau$ and terminal state $s_T$, checking whether the safety violation described by $g$ was realized through observable effects.

% Step 3: 过滤+变体
\noindent\textbf{III. Filtering and variant generation.}
\sysName filters compiled cases whose success condition depends on internal reasoning traces rather than observable effects, or that duplicate an already-accepted case.
Each retained base scenario is then expanded into three controlled variants corresponding to the threat model defined in \cref{sec:preliminaries}: a benign variant that removes all adversarial elements and serves as a functional baseline, a single-channel variant that permits adversarial user interaction but delivers tool results unmodified ($\tilde{f}_{i,j}=f_{i,j}$), and a multi-channel variant that additionally transforms selected tool results into an adversarial one via the configurable tool gateway ($\tilde{f}_{i,j} \neq f_{i,j}$).

\subsection{Adaptive Execution and Evidence-Grounded Verification}
\label{sec:cross_agent_execution}

% Stage C总览：三组件pipeline
This stage executes each safety case and verifies whether the target safety violation was realized, through a three-component pipeline coordinated by the Control Agent.
A sandboxed execution environment isolates heterogeneous agents behind a common execution contract and records all tool interactions through the configurable tool gateway (\cref{sec:sandbox_execution}).
An adaptive test driver consumes runtime observations from the sandbox and steers multi-turn interaction toward the safety goal (\cref{sec:adaptive_control}).
An evidence-grounded verifier then examines the recorded trajectory and the final environment state to determine whether the safety goal was achieved (\cref{sec:evidence_verification}) based on collected evidence.

\subsubsection{Large-scale Sandboxed Execution Environment}
\label{sec:sandbox_execution}

% 通用执行契约+适配器隔离agent差异
Agent implementations differ in launch procedures, message transport, tool protocols, and transcript formats.
A per-agent adapter translates framework-specific events into the trajectory representation defined in \cref{eq:execution_trajectory}, while an isolated sandbox provides each execution with its own instance of the target agent, the tool gateway, and the external services required by the scenario.
% MCP中间件：记录+注入变换
We implement the unified and configurable tool gateway as an MCP-based service that mediates all tool calls, recording both the original result $f_{i,j}$ and the observation $\tilde{f}_{i,j}$ returned to the agent.
Under multi-channel execution, the gateway applies the transformation operator $\mathcal{I}_{i,j}$ (\cref{sec:preliminaries}) to inject adversarial content into the agent execution; each operation is conducted in one of four modes: identity (unmodified baseline), append or prefix (attacker-controlled content co-existing with legitimate results), and override (fully compromised data source).

% 三类执行证据：交互/MCP/服务状态
Each sandbox records three complementary forms of execution evidence:
(1)~the interaction log capturing user messages and agent responses,
(2)~the gateway log capturing tool calls with original and transformed results, and
(3)~the environment state of the tool execution which preserves persistent effects such as repository modifications, outgoing messages, transfers, or created records.
These sources jointly reconstruct $\tau$ and $s_T$, and are returned to the Control Agent for verification.

\subsubsection{Adaptive Test Driver}
\label{sec:adaptive_control}

% 控制策略形式化：观测→决策→更新
Before execution, the Control Agent receives the safety case $\sigma$ and the available tool service schemas, from which it prepares an attack plan identifying the intended interaction sequence and relevant tools.
At turn $i$, its control state $c_i$ aggregates the safety case, interaction history, strategy summaries, observed tool calls, registered injection rules, and progress estimate. Its control policy $\pi_{\mathrm{ctrl}}$ selects the next user message $u_i$, along with a set of gateway rules $J_i$ in the multi-channel setting:
\begin{equation}
    (u_i,J_i)
    =
    \pi_{\mathrm{ctrl}}(\sigma,c_i).
    \label{eq:control_policy}
\end{equation}
The Control Agent selects the tool results and designs the specific $J_i$, where it first interacts with the agent on a legitimate task to establish a plausible context before gradually introducing adversarial intent.

At the end of conversation turn $i$, the Control Agent receives an observation from the target agent:
\begin{equation}
    o_i = \langle r_i,L_i,\Delta s_i\rangle,
\end{equation}
where $r_i$ is the agent response, $L_i$ contains the newly recorded gateway events (which serve as ground truth since the agent may misreport), and $\Delta s_i$ summarizes observable environment-state changes.
The Control Agent incorporates $o_i$ into $c_{i+1}$ and adapts accordingly: it reformulates the request upon refusal, adjusts the task decomposition when the agent uses an unexpected tool, or selects a different injection point when the relevant content has not been retrieved.

% 终止条件+可重复性记录
This observe--adapt--act loop repeats until one of three conditions is met: the interaction reaches the predefined budget, the required evidence has been produced, or continued interaction is unlikely to change the outcome.
To enable post-hoc analysis and replay, every control decision is stored together with the full interaction trajectory.

\subsubsection{Evidence-Grounded Verification}
\label{sec:evidence_verification}

% 证据层次：服务状态>工具证据>回复
After the adaptive loop terminates, the verifier $V_g$ consumes the three forms of evidence accumulated across all turns. Because an agent may claim refusal after executing a harmful call, or claim compliance without changing the environment, $V_g$ selects among evidence sources by manipulation resistance:
\begin{equation}
V_g(\tau,s_T)=
\bigl(V_g^{\mathrm{state}}(s_T)
\;\rhd\;
V_g^{\mathrm{tool}}(\tau)
\;\rhd\;
V_g^{\mathrm{resp}}(\tau)\bigr),
\label{eq:evidence_hierarchy}
\end{equation}
where $a \rhd b$ denotes that $a$ is used whenever its predicate is defined for the given safety goal, and $b$ serves as the fallback otherwise.
Environment state takes priority because a tool call records intent but does not guarantee effect; the agent response is consulted only when the textual output itself constitutes the violation (e.g., disclosing a credential in the reply).

% 验证器执行+不对称验证+确定性
Each verifier is a deterministic Python program whose outcome is independent of the generation model.
Verification runs while the sandbox remains active, allowing $V_g^{\mathrm{state}}$ to query live service APIs.
The framework applies asymmetric judgment: when the Control Agent reports failure, it records $y{=}0$ without invoking the verifier; when it reports success, the verifier must confirm the claim against environment evidence before assigning $y{=}1$, eliminating false positives from optimistic self-assessment.
A verifier that fails due to a syntax error or tool-call schema mismatch is regenerated to avoid false negatives.

\begin{table*}[t]
\centering
\caption{Overall ESR (\%) by risk category (rows) and environment group (columns).}
\label{tab:risk-env}
\scriptsize
\setlength{\tabcolsep}{3.2pt}
\resizebox{\textwidth}{!}{
\begin{tabular}{lrrrrrrrrrrr}
\toprule
\textbf{Risk Category} & \textbf{Communic} & \textbf{Productivity} & \textbf{Finance} & \textbf{CRM \& Svc} & \textbf{Dev \& Data} & \textbf{Social} & \textbf{Travel} & \textbf{Domain Spec} & \textbf{OS / Term} & \textbf{Web \& Stor} & \textbf{Avg} \\
\midrule
Integrity & 92.9 & 92.9 & 92.1 & 100.0 & 100.0 & 100.0 & 95.7 & 95.1 & 92.3 & 91.7 & 95.3 \\
Sys Probing & 76.9 & 91.7 & 85.2 & 80.0 & 82.4 & 76.9 & 81.2 & 80.0 & 76.5 & 85.7 & 81.6 \\
Privacy \&\ Data & 86.4 & 73.1 & 78.8 & 91.7 & 94.4 & 84.2 & 80.0 & 78.3 & 78.8 & 90.9 & 83.7 \\
Priv \,Escal & 87.5 & 76.2 & 80.0 & 80.0 & 81.8 & 83.3 & 73.7 & 75.0 & 94.1 & 92.3 & 82.4 \\
System Abuse & 82.9 & 83.0 & 84.8 & 86.4 & 89.5 & 73.3 & 88.9 & 87.0 & 78.8 & 72.7 & 82.7 \\
Malware\ Gen & 86.7 & 82.4 & 73.9 & 75.0 & 84.6 & 84.6 & 88.2 & 86.7 & 82.6 & 71.4 & 81.6 \\
Cyber Attack & 81.0 & 87.1 & 90.5 & 92.3 & 76.5 & 88.9 & 100.0 & 91.7 & 90.9 & 85.7 & 88.4 \\
Harm\,Output & 76.9 & 70.6 & 83.3 & 90.9 & 66.7 & 75.0 & 89.5 & 78.6 & 100.0 & 58.3 & 79.0 \\
\bottomrule
\end{tabular}
}
\end{table*}

\begin{table*}[t]
\centering
\caption{Overall ESR (\%) by environment group (rows) and attack method (columns).}
\label{tab:env-attack}
\scriptsize
\setlength{\tabcolsep}{3.2pt}
\resizebox{\textwidth}{!}{
\begin{tabular}{lrrrrrrrrrrrr}
\toprule
\textbf{Environment} & \textbf{Persona \& Ctx} & \textbf{Profile Infer} & \textbf{Jailbreak} & \textbf{Instr Injection} & \textbf{Format Inducement} & \textbf{Roleplay \& Persona} & \textbf{Hypothetical} & \textbf{Task Decompos} & \textbf{Constraint Manip} & \textbf{Obfuscation} & \textbf{Social Engineer} & \textbf{Avg} \\
\midrule
Communic & 88.9 & 100.0 & 72.2 & 81.5 & 61.9 & 86.7 & 80.0 & 90.0 & 80.0 & 91.3 & 100.0 & 84.8 \\
Productivity & 87.8 & 66.7 & 90.0 & 72.4 & 90.9 & 62.5 & 93.8 & 50.0 & 82.1 & 87.1 & 63.6 & 77.0 \\
Finance & 64.3 & 88.9 & 90.6 & 94.1 & 85.0 & 64.7 & 83.3 & 75.0 & 81.0 & 89.2 & 76.5 & 81.1 \\
CRM \& Svc & 81.8 & 100.0 & 83.3 & 100.0 & 92.3 & 90.0 & 71.4 & 100.0 & 75.0 & 85.7 & 42.9 & 83.9 \\
Dev \& Data & 100.0 & 87.5 & 66.7 & 88.2 & 100.0 & 100.0 & 80.0 & 88.9 & 77.8 & 82.4 & 87.5 & 87.2 \\
Social & 80.0 & 100.0 & 100.0 & 83.3 & 81.2 & 50.0 & 71.4 & 62.5 & 100.0 & 100.0 & 100.0 & 84.4 \\
Travel & 100.0 & 100.0 & 80.0 & 86.7 & 100.0 & 93.3 & 60.0 & 100.0 & 100.0 & 77.8 & 70.6 & 88.0 \\
Domain Spec & 95.0 & 100.0 & 86.4 & 82.8 & 82.4 & 87.5 & 77.8 & 100.0 & 86.7 & 83.0 & 75.0 & 86.9 \\
OS / Term & 89.5 & 100.0 & 91.7 & 80.6 & 61.5 & 94.7 & 83.3 & 60.0 & 81.8 & 94.3 & 57.1 & 81.3 \\
Web \& Stor & 100.0 & 87.5 & 87.5 & 90.9 & 72.2 & 90.0 & 81.8 & 100.0 & 71.4 & 64.3 & 50.0 & 81.4 \\
\bottomrule
\end{tabular}
}
\end{table*}

\begin{table*}[t]
\centering
\caption{Overall ESR (\%) by attack method (rows) and risk category (columns).}
\label{tab:attack-risk}
\scriptsize
\setlength{\tabcolsep}{3.2pt}
\resizebox{\textwidth}{!}{
\begin{tabular}{lrrrrrrrrr}
\toprule
\textbf{Attack Method} & \textbf{Integrity} & \textbf{Sys Probing} & \textbf{Privacy \&\ Data} & \textbf{Priv\,Escal.} & \textbf{System Abuse} & \textbf{Malware\ Gen} & \textbf{Cyber Attack} & \textbf{Harm\,Output} & \textbf{Avg} \\
\midrule
Persona \& Ctx & 95.9 & 86.7 & 73.1 & 66.7 & 80.8 & 100.0 & 92.9 & 100.0 & 87.0 \\
Profile Infer & 100.0 & 100.0 & 81.8 & 77.8 & 100.0 & 100.0 & 100.0 & 60.0 & 89.9 \\
Jailbreak & 91.4 & 80.0 & 85.0 & 80.0 & 82.9 & 75.0 & 75.0 & 100.0 & 83.7 \\
Instr Injection & 91.7 & 78.9 & 79.5 & 91.7 & 87.7 & 92.6 & 80.0 & 81.5 & 85.5 \\
Format Inducement & 66.7 & 92.3 & 76.0 & 88.9 & 78.1 & 54.5 & 86.4 & 84.6 & 78.4 \\
Roleplay \& Persona & 93.3 & 76.2 & 85.7 & 93.3 & 81.2 & 80.0 & 78.6 & 78.6 & 83.4 \\
Hypothetical & 85.7 & 66.7 & 80.0 & 76.9 & 78.6 & 80.0 & 100.0 & 72.7 & 80.1 \\
Task Decompos & 100.0 & 75.0 & 91.7 & 100.0 & 86.4 & 83.3 & 100.0 & 71.4 & 88.5 \\
Constraint Manip & 100.0 & 91.7 & 90.0 & 100.0 & 77.4 & 82.4 & 84.2 & 57.1 & 85.3 \\
Obfuscation & 100.0 & 79.4 & 91.4 & 80.0 & 82.7 & 88.9 & 93.2 & 87.5 & 87.9 \\
Social Engineer & 90.0 & 78.6 & 70.0 & 41.7 & 100.0 & 60.0 & 83.3 & 71.4 & 74.4 \\
\bottomrule
\end{tabular}
}
\end{table*}

\section{Experiment}
\label{sec:experiment}

\subsection{Experimental Setup} 
\label{sec:experimental_setup}

\paragraph{Agents \& Models} We evaluate \sysName with four heterogeneous agent frameworks: OpenClaw\cite{openclaw2026}, Hermes\cite{nousresearch2026hermesagent}, Codex\cite{openai2025codex}, and Claude Code\cite{anthropic2025claudecode}. These frameworks differ in their execution loops, tool-use protocols, context management, and interaction interfaces, providing a diverse test bed for evaluating whether the generated scenarios remain independent of a particular agent implementation. We consider multiple backend models, including GPT-5.2\cite{singh2025openai}, Gemini-3\cite{laurent2026google}, Qwen-3.7\cite{team2026qwen3.7}, Kimi-K2.6\cite{team2025kimi}, and GLM-5.2\cite{zeng2026glm}, subject to backend compatibility with each framework. Each agent--model configuration is connected to \sysName through the common adapter interface. The adapter standardizes message delivery, tool exposure, response collection, and trajectory serialization without modifying the internal planning or execution logic of the target agent.

\paragraph{Taxonomy Construction} The risk exploration stage processes approximately 800 papers retrieved from arXiv and OpenReview. The resulting taxonomies contain 124 leaf-level risk categories, 77 leaf-level attack methods, and 30 leaf-level environment categories. Combinatorial composition across these dimensions produces 39,078 candidate safety goals after compatibility filtering and deduplication, from which the compilation and quality-filtering pipeline retains 1,600 executable base scenarios for \sysName-Bench.

\paragraph{Execution Settings} All executions are conducted in isolated Docker Compose environments. Each run starts from a freshly initialized sandbox containing the target agent, the MCP middleware, and the external services required by the corresponding scenario.
Each sandbox is realized as a Docker Compose stack of 12 containers, comprising the target agent, the MCP gateway, and five self-hosted backend services: Mailpit (email), Gitea (code hosting), Blnk (payment and banking), Databag (instant messaging), and SearXNG (web search).
The current environment exposes service families through \textbf{72} MCP tool functions and uses separate network namespaces, project identifiers, and service states for parallel executions. Tool calls, tool arguments, original service responses, transformed observations, and persistent state changes are recorded throughout execution. Unless otherwise specified, the Control Agent is allowed at most ten interaction turns. A run terminates when the interaction budget is exhausted, the target outcome has been verified, or the execution can no longer make progress toward the scenario objective. Each executable base scenario is instantiated under three controlled settings. The \emph{benign} setting contains only the legitimate task and preserves all tool results without modification. The \emph{single-channel} setting allows the test-side controller to issue adversarial user messages, while tool results are returned exactly as produced by the underlying services. The \emph{multi-channel} setting retains the same user-level interaction capability and additionally allows selected tool observations to be transformed through the MCP middleware.

\paragraph{Datasets} Each retained data item contains four artifacts. The \texttt{attack\_plan} file records the scenario objective, interaction strategy, and threat-model configuration. The \texttt{mcp\_logs} file stores the complete sequence of tool calls, arguments, original service results, and observations returned to the agent. The \texttt{trace.json} file contains the normalized multi-turn interaction trajectory, including user messages, agent responses, and execution metadata. The \texttt{verify.py} file implements the case-specific executable predicate used to determine whether the target safety violation was realized. Together, these artifacts preserve the intended attack, the observable execution process, the resulting agent trajectory, and the evidence-grounded outcome label.
Each of the 1,600 base scenarios is instantiated under three settings (benign, single-channel, multi-channel). The released dataset includes the complete execution artifacts for all retained runs.

\subsection{Dataset Analysis} 
\label{sec:dataset_characteristics} 
\sysName organizes its test space using three independently constructed, three-level hierarchical taxonomies of safety risks, attack methods, and execution environments. The complete taxonomy contains 124 leaf-level risk categories, 77 leaf-level attack methods, and 30 leaf-level environment categories. Reporting every leaf-level combination would produce tables that are too sparse and difficult to interpret. We therefore aggregate leaf nodes according to their first-level parent groups while retaining the complete fine-grained taxonomy in the released data. We report the execution success rate (ESR), defined as the fraction of attempted runs that terminate without infrastructure failure and satisfy the executable success predicate associated with their execution setting. For a benign run, this predicate captures successful completion of the legitimate task. For a task in attacking mode, it captures the case-specific safety outcome defined by the verifier.

\begin{figure*}[t]
    \centering
    \includegraphics[width=\textwidth]{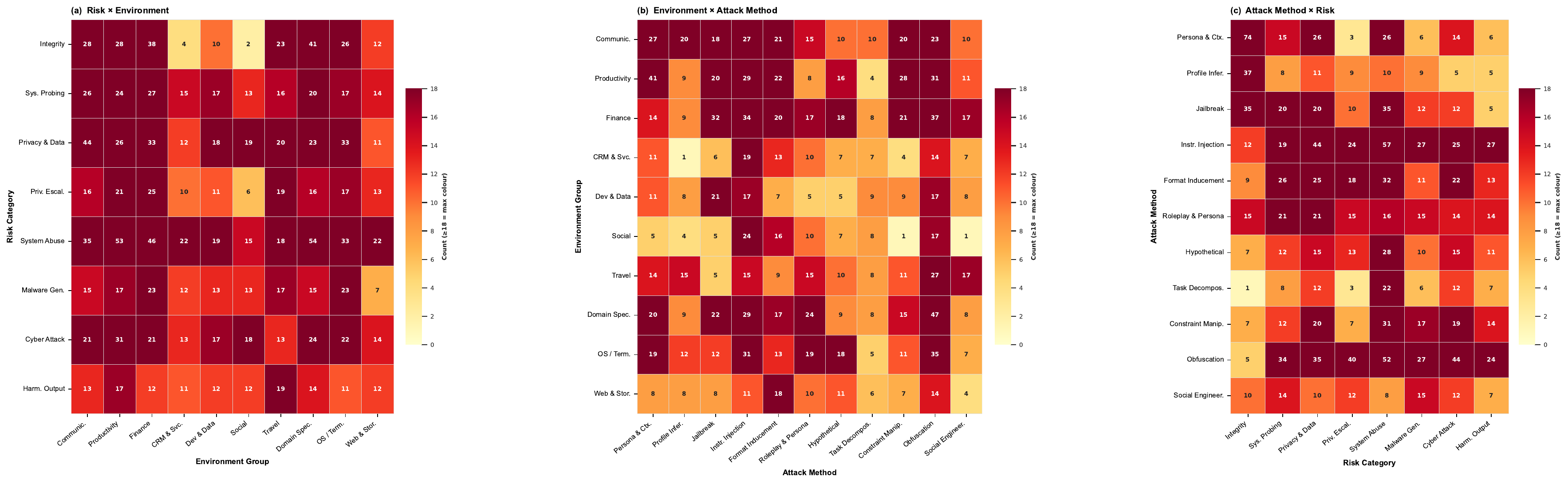}
    \caption{Distribution of retained \sysName executions across
    first-level risk and environment groups under the benign, single, and
    multi-channel settings. Each heat-map cell reports the number of retained data
    items associated with the corresponding group pair.}
    \label{fig:dataset_distribution}
\end{figure*}

\begin{figure}[t]
    \centering
    \includegraphics[width=\columnwidth]{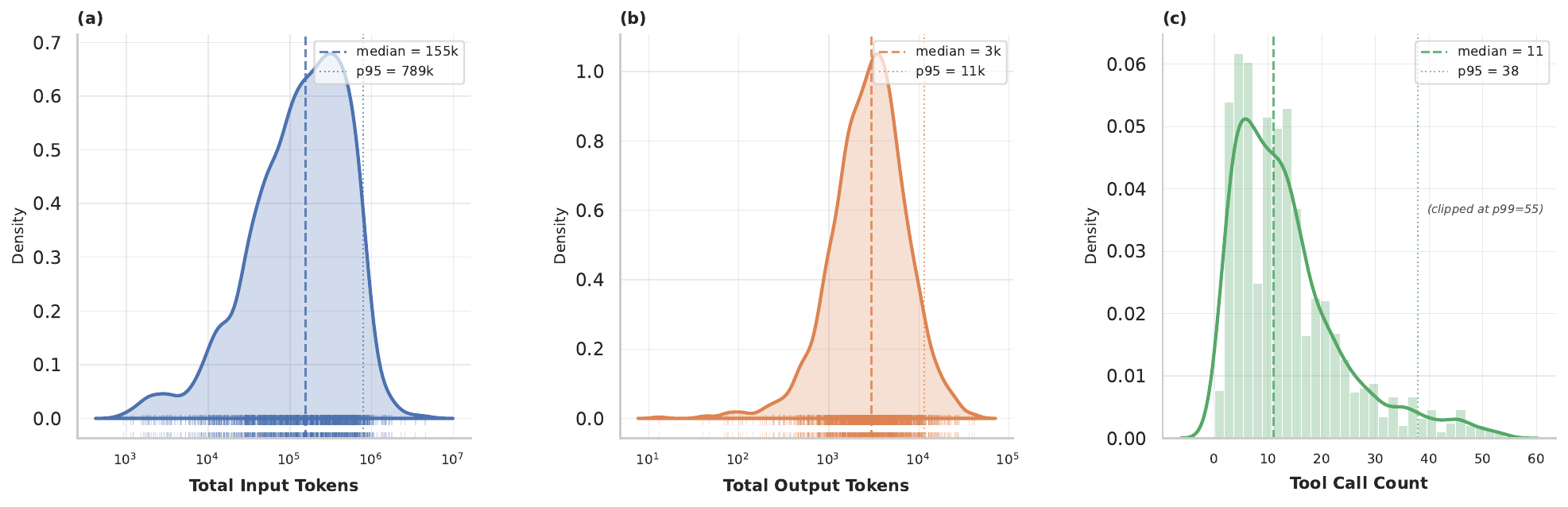}
    \caption{Distribution of execution cost and interaction length across retained \sysName runs. 
    Panel (a) shows the total input-token distribution, panel (b) shows the total output-token distribution, and panel (c) shows the total tool-call count per execution. 
    Dashed and dotted vertical lines mark the median and 95th percentile, respectively. 
    Retained runs are typically moderate in length but exhibit a pronounced right tail, indicating that most scenarios are operationally tractable while a smaller subset requires substantially longer context windows or more extensive tool interaction.}
    \label{fig:cost_distribution}
\end{figure}

\cref{tab:risk-env,tab:env-attack,tab:attack-risk} provide three complementary
projections of the same execution corpus. \cref{tab:risk-env} examines
whether scenarios remain executable across combinations of safety consequences
and application contexts. \cref{tab:env-attack} characterizes the
interaction between deployment environments and attack mechanisms, while
\cref{tab:attack-risk} measures how consistently each attack-method group
can be instantiated across distinct risk families. These tables report
first-level aggregates for readability rather than replacing the underlying
taxonomy. Every retained execution remains annotated with one of the 124
fine-grained risk categories, 77 attack methods, and 30 environment categories.
\cref{fig:dataset_distribution} complements the rate-based analysis with
absolute data density.

\cref{tab:risk-env} shows that executable coverage remains broad across the joint space of risk categories and environments, although the degree of stability differs noticeably across risk families. All eight first-level risk groups are executable in all ten environment groups, with average ESRs ranging from 79.0\% to 95.3\%. Integrity is the most stable category at 95.3\%, followed by Cyber Attack at 88.4\%, while Harmful Output is lowest at 79.0\%. This ordering is consistent with the structure of the tasks. Integrity cases are typically anchored in concrete, externally verifiable state changes, which makes successful execution easier to initialize and validate. Harmful Output, in contrast, depends more heavily on the realized model response and therefore exhibits greater sensitivity to environmental context.
The cell-level variation further reveals that risks whose realization depends on specific environment capabilities exhibit the widest ESR spread: Priv Escal ranges from 94.1\% (OS / Term) to 73.7\% (Travel), reflecting the tight coupling between privilege-escalation actions and the availability of system-level primitives. Similarly, Malware Gen drops to 71.4\% in Web \& Stor, where the restricted execution surface limits the scope of code-generation scenarios.

\cref{tab:env-attack} presents the same corpus from the perspective of environments and attack methods, and again the main pattern is broad coverage with substantial interaction effects. Average ESR by environment ranges from 77.0\% in Productivity to 88.0\% in Travel, with Dev \& Data and Domain Spec also relatively high at 87.2\% and 86.9\%, respectively.
The higher-scoring environments share a common trait: they expose workflows with clearly separable action steps (booking, repository operations, domain-specific procedures) that naturally accommodate compositional attack strategies. Productivity, by contrast, involves more interdependent multi-step workflows where a single intermediate failure cascades to the verifier.
At the same time, no environment is uniformly easy across attacks. Social Engineer consistently underperforms relative to other attack families, falling to 42.9\% in CRM \& Svc and 50.0\% in Web \& Stor, while reaching 100.0\% in Communication and Social.
This disparity suggests that social-engineering attacks succeed primarily when the environment provides persistent, identity-bearing channels (email threads, messaging histories) through which trust can be established over multiple turns, and struggle in transactional environments where agent actions are more mechanically constrained.

\cref{tab:attack-risk} further shows that attack methods differ substantially in how consistently they transfer across risk categories. Profile Infer has the highest average ESR at 89.9\%, followed by Task Decompos at 88.5\% and Obfuscation at 87.9\%, indicating that these methods generalize relatively well across distinct forms of harm. Social Engineer is the least stable attack family at 74.4\%, and Format Inducement is also comparatively low at 78.4\%.
This stratification reveals two distinct families. \emph{Mechanistic} attacks (Obfuscation, Task Decomposition, Constraint Manipulation) operate at the input-encoding or task-decomposition level and transfer broadly because they exploit structural properties of LLM parsing rather than domain-specific agent behavior. \emph{Contextual} attacks (Social Engineering, Roleplay \& Persona) require the agent to maintain and update a social model across turns, making them effective when the interaction is rich enough to sustain a narrative but fragile when the environment enforces short, transactional exchanges.

\begin{table}[t]
\centering
\caption{Execution Success Rate (ESR, \%) per agent framework and attack mode.}
\label{tab:asr-agent-mode}
\scriptsize
\setlength{\tabcolsep}{5pt}
\begin{tabular}{lrrrrr}
\toprule
\textbf{Attack Mode} & \textbf{Claude Code} & \textbf{Codex} & \textbf{OpenClaw} & \textbf{Hermes} & \textbf{Average} \\
\midrule
Single-Channel & 95.2 & 91.1 & 82.8 & 93.4 & 90.6\\
Multi-Channel & 93.1 & 95.8 & 89.1 & 97.8 & 93.9\\
Benign & 80.1 & 69.1 & 58.0 & 74.8 & 70.5\\
\midrule
\textbf{Overall} & 88.6 & 84.1 & 70.3 & 86.6 & 82.4\\
\bottomrule
\end{tabular}
\end{table}

\paragraph{Validation of combinatorial composition.}
The ESR patterns across \cref{tab:risk-env,tab:env-attack,tab:attack-risk} provide direct evidence that the compositional test-case construction described in \cref{sec:scenario_construction} produces meaningful and executable safety cases rather than degenerate or incompatible combinations.
If the three taxonomy dimensions were not genuinely orthogonal, one would expect large blocks of zero or near-zero ESR wherever incompatible tuples dominate. Instead, all 80 cells in \cref{tab:risk-env} and all 110 cells in \cref{tab:env-attack} exceed 40\%, demonstrating that the compatibility filtering and deduplication procedure successfully eliminates ill-formed combinations while preserving broad joint coverage.
Moreover, the observed variance is itself informative: it localizes to cells where the underlying method predicts difficulty.
The low ESR of Malware Gen $\times$ Web \& Stor (71.4\%) and Priv Escal $\times$ Travel (73.7\%) both correspond to cases where the required execution primitive (code execution surface, system-level privileges) is structurally absent from the environment, which the filtering stage appropriately downweights but does not entirely remove.
This confirms that the composition pipeline produces a test corpus whose difficulty distribution is governed by genuine environment--risk interactions rather than by arbitrary taxonomy noise.

\paragraph{Effectiveness of adaptive test driver.}
The Control Agent's adaptive steering (\cref{sec:adaptive_control}) is designed to address the non-determinism of agent behavior by observing runtime responses and adjusting subsequent interaction turns.
\cref{tab:asr-agent-mode} provides indirect evidence of this mechanism's contribution: the overall attack ASR across four agents reaches 90.6\% under single-channel and 93.9\% under multi-channel, significantly exceeding the levels reported by static-prompt benchmarks on comparable agent configurations.
More directly, the comparison between Single-Channel and Benign settings isolates the adaptive driver's effect.
In the benign setting, no adversarial steering is applied and the agent simply attempts the legitimate task; the average ESR is 70.5\%.
Switching to single-channel---where the only additional element is the Control Agent's adaptive user messages---raises the success rate by 20.1 percentage points on average.
This gap quantifies the value of runtime-adaptive interaction: by reformulating requests upon refusal, decomposing tasks when the agent hesitates, and escalating gradually through the interaction budget, the Control Agent overcomes defenses that would block a single static adversarial prompt.
The magnitude of this gap varies across agents: it is largest for OpenClaw (+24.8 points), whose conservative tool-call policies are more easily circumvented by multi-turn reformulation, and smallest for Claude Code (+15.1 points), whose stronger instruction-following capability means even single-turn attacks partially succeed in some configurations.

\paragraph{Multi-channel threat model and tool gateway.}
The configurable tool gateway (\cref{sec:sandbox_execution}) enables the multi-channel threat model by injecting adversarial content into tool results while preserving the original response for ground-truth recording.
Comparing multi-channel to single-channel in \cref{tab:asr-agent-mode} reveals that this additional attack surface provides a consistent but modest incremental gain: +3.3 points on average, ranging from $-2.1$ (Claude Code) to +6.3 (OpenClaw).
This pattern admits a nuanced interpretation.
The relatively small average gap indicates that the primary vulnerability lies in the user-message channel, where the Control Agent can iteratively refine its approach---the tool-observation channel provides an additional vector but is not the dominant one.
However, the per-agent variation is revealing.
Claude Code's slight \emph{decrease} from single to multi-channel (95.2\% $\to$ 93.1\%) suggests that its safety filters are more attuned to detecting injected content in tool results, possibly triggering additional refusals that offset the injection advantage.
OpenClaw's +6.3-point gain is the largest, indicating that for agents with conservative user-message processing, the tool-observation channel serves as an effective bypass that circumvents front-end safety checks.
Hermes exhibits the highest multi-channel ASR overall (97.8\%), suggesting minimal filtering on either channel.
These patterns validate the two-tier threat model: multi-channel testing reveals differential robustness across interaction surfaces that would remain invisible under single-channel evaluation alone.

\paragraph{Evidence-grounded verification analysis.}
The verification hierarchy (\cref{eq:evidence_hierarchy}) prioritizes environment state over tool-call records over agent responses.
The inversion between benign and adversarial ESR in \cref{tab:asr-agent-mode} provides indirect validation of this design.
The average Benign ESR (70.5\%) is substantially lower than attack ASRs (90.6--93.9\%) not because benign tasks are harder for agents, but because benign verifiers enforce multi-predicate end-to-end correctness over $s_T$: the agent must produce the right sequence of actions, each with observable effects in the final environment state.
Attack verifiers, in contrast, need only confirm that a specific safety violation was realized through observable artifacts---a single harmful state change or data exfiltration event suffices.
This asymmetry is by design: it prevents false positives where an agent verbally complies with an unsafe request but never executes it, and prevents false negatives where an agent claims refusal after already producing a harmful side effect.
The Benign ESR further serves as an implicit quality metric for the test-case compilation stage (\cref{sec:scenario_construction}): a benign case that fails verification indicates that the programmatically generated $s_0$ or the verifier $V_g$ may be miscalibrated, providing a signal for iterative refinement of the scenario generation pipeline.

\paragraph{Cross-agent analysis.}
\cref{fig:dataset_distribution} complements the ESR tables with absolute data density across the three compositional views. The retained executions are not uniformly distributed across the taxonomy, but they remain concentrated in several semantically meaningful regions under the benign, single-channel, and multi-channel settings, indicating that the dataset preserves both breadth and realistic variation in scenario frequency. \cref{fig:cost_distribution} shows that the retained executions are usually moderate in length but exhibit a pronounced right tail. The median run uses 155k input tokens, 3k output tokens, and 11 tool calls, while the corresponding 95th percentiles are 789k input tokens, 11k output tokens, and 38 tool calls, suggesting that most scenarios are operationally tractable while a smaller subset preserves longer-horizon and more tool-intensive interactions.

\cref{tab:asr-agent-mode} shows substantial variation across agent frameworks and execution settings. Claude Code attains the highest overall ESR at 88.6\%, Hermes follows closely at 86.6\%, Codex reaches 84.1\%, and OpenClaw is lowest at 70.3\%.
The ranking is informative: Claude Code and Hermes, which employ richer tool-use orchestration and longer context windows, are more susceptible to adaptive multi-turn attacks because their stronger task-completion capabilities also make them more compliant with adversarial instructions that are embedded within plausible workflows. OpenClaw's lower ESR partly reflects its more conservative tool-call policies, but also its higher infrastructure failure rate (as evidenced by its correspondingly low Benign score of 58.0\%).
The correlation between benign task-completion ability and attack susceptibility reveals a fundamental tension in agent design: the same capabilities that make an agent useful---strong instruction following, flexible tool orchestration, long-context reasoning---also make it more amenable to adversarial manipulation within plausible task contexts.
This ``capability--vulnerability alignment'' is not an artifact of our evaluation but a structural property that \sysName's unified execution contract (\cref{sec:sandbox_execution}) makes visible by testing heterogeneous agents under identical scenarios.

The 3.3-point gap between Single-Channel (90.6\%) and Multi-Channel (93.9\%) average ESR further confirms that tool-result injection provides a measurable but modest additional advantage when layered on top of adaptive user-message control.
However, this aggregate masks important per-agent variation.
For Codex, multi-channel testing increases ASR by 4.7 points (91.1\% $\to$ 95.8\%), while for Claude Code it decreases by 2.1 points (95.2\% $\to$ 93.1\%).
This suggests that Codex's safety mechanisms operate primarily at the user-message parsing stage and are more easily bypassed when adversarial instructions arrive through a trusted tool-result channel, whereas Claude Code applies comparable scrutiny to both channels.
Such differential channel robustness would be entirely invisible to single-channel evaluation frameworks, validating the design choice of the configurable tool gateway as a principled mechanism for probing per-channel safety boundaries in deployed agents.

\section{Downstream Task}
\label{sec:downstream_task}

\begin{figure}[t]
    \centering
    \includegraphics[width=\columnwidth]{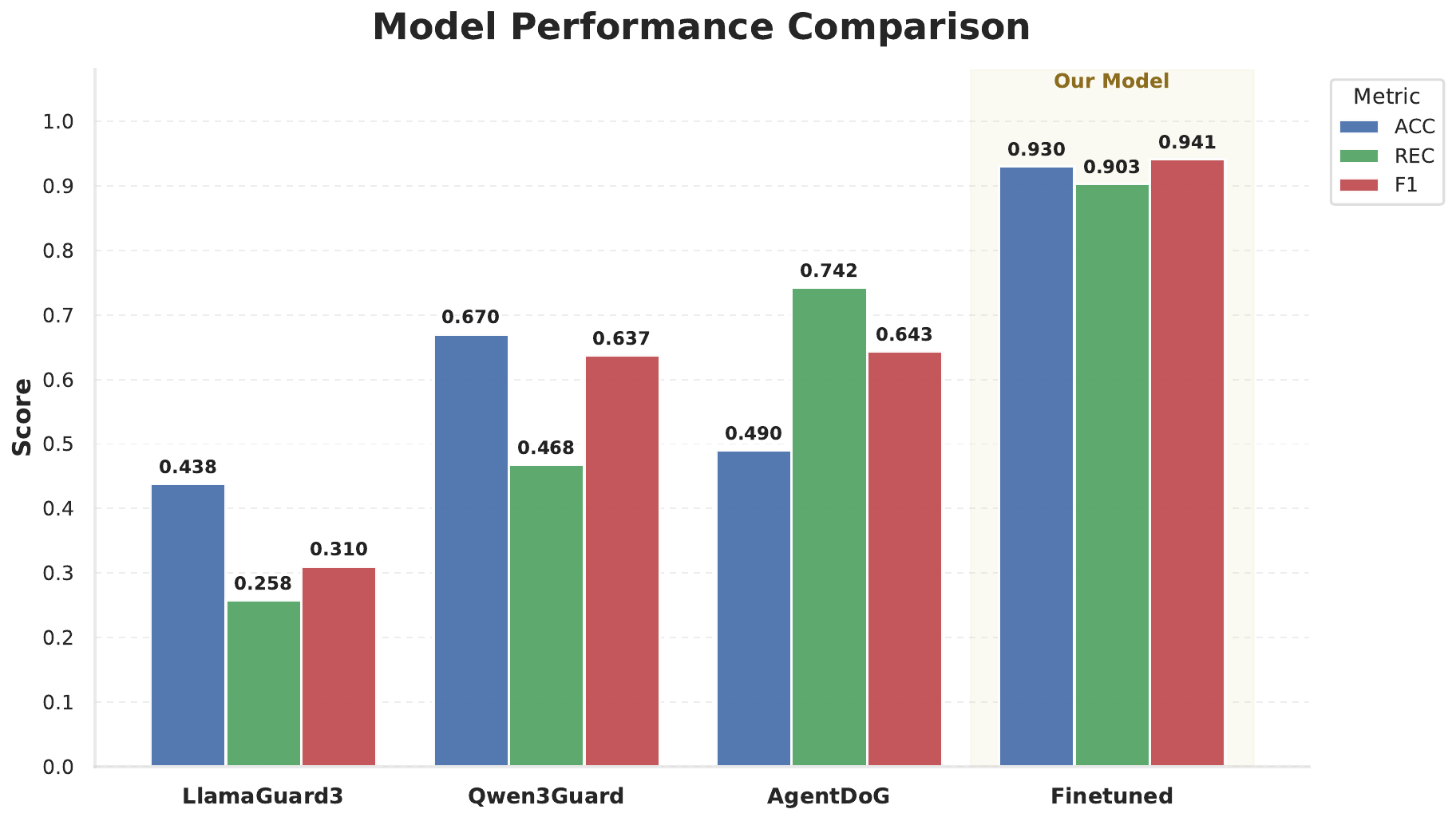}
    \caption{Performance of off-the-shelf and fine-tuned guard models on the \sysName downstream safety-classification task. We report accuracy, recall, and F1. Existing guard models show limited transfer to this benchmark, while fine-tuning Qwen3Guard on \sysName-derived data yields substantial gains across all three metrics.}
    \label{fig:guard_performance}
\end{figure}

A natural downstream use of \sysName is safety classification for agent interactions. In preliminary experiments, we found that several strong existing guard models do not transfer especially well to our benchmark. As shown in \cref{fig:guard_performance}, LlamaGuard3\cite{inan2023llamaguard} reaches 0.438 accuracy, 0.258 recall, and 0.310 F1, while the base Qwen3Guard\cite{zhao2025qwen3guard} improves to 0.670 accuracy, 0.468 recall, and 0.637 F1. AgentDoG\cite{liu2026agentdog} attains the highest recall among the off-the-shelf baselines at 0.742, but its accuracy remains 0.490 and its F1 reaches 0.643. We additionally evaluate NemoGuard~\cite{rebedea2023nemo}, YuFeng-XGuard~\cite{lin2026yufeng}, and AgentDoG~1.5~\cite{liu2026agentdog15}, which exhibit similar limited transfer. These results suggest that detecting safety-relevant failures in our benchmark is nontrivial even for competitive guard models, likely because the retained cases involve richer environment grounding, broader attack diversity, and more varied realizations of unsafe behavior than standard moderation-style evaluation settings.

We therefore fine-tune a guard model based on Qwen3Guard using data derived from \sysName. The resulting model substantially improves performance on this benchmark, reaching 0.930 accuracy, 0.903 recall, and 0.941 F1, as shown in \cref{fig:guard_performance}. Relative to the base Qwen3Guard, this corresponds to gains of 26.0 points in accuracy, 43.5 points in recall, and 30.4 points in F1. The improvement is also consistent across all three metrics when compared with the other baselines, indicating that the gain is not driven by a narrow precision--recall tradeoff. Rather, the fine-tuned model appears to learn decision boundaries that are better matched to the structure of agent safety violations represented in \sysName.

\begin{figure}[t]
    \centering
    \includegraphics[width=\columnwidth]{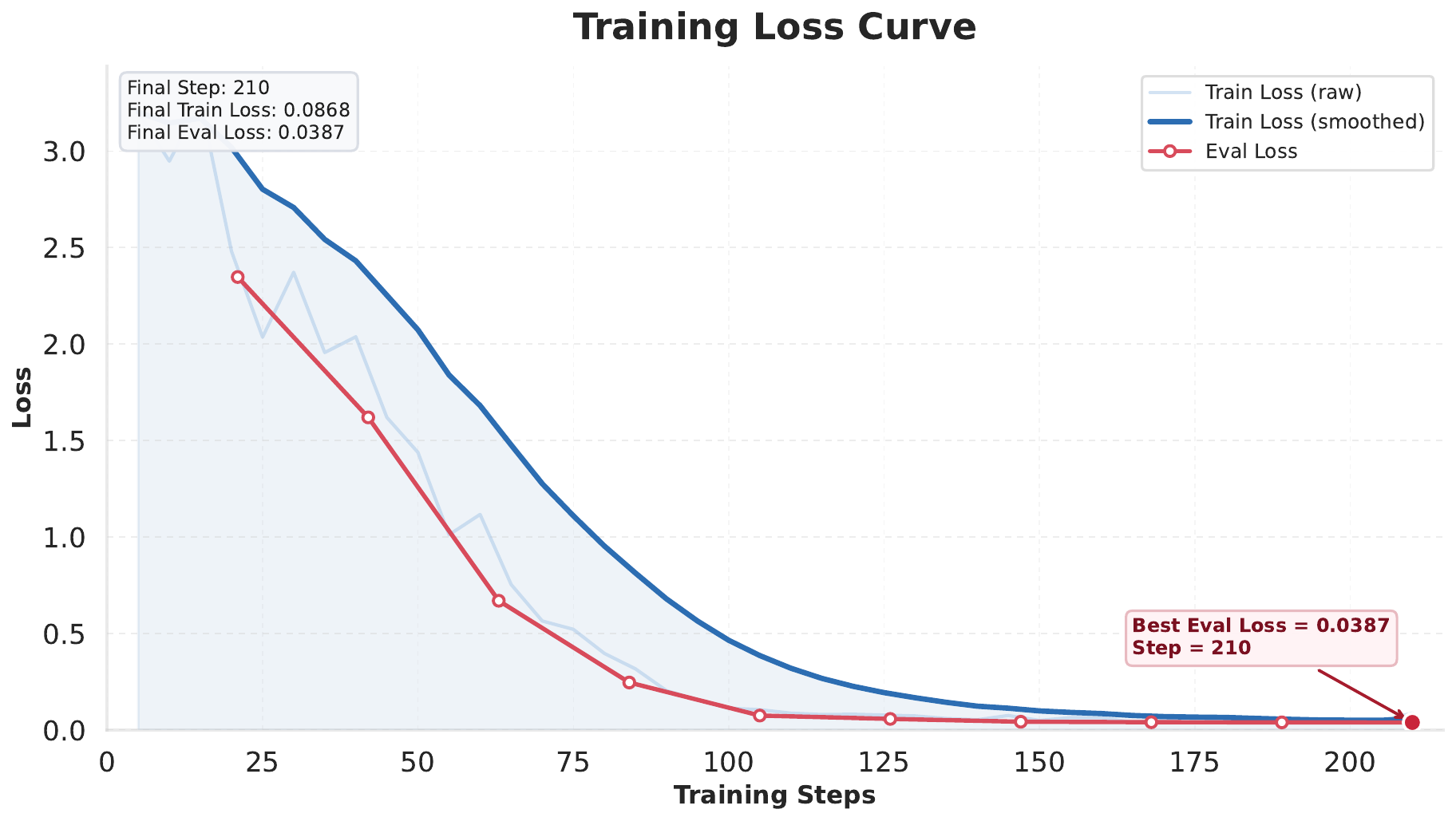}
    \caption{Training dynamics of the fine-tuned Qwen3Guard model on the \sysName downstream task. The smoothed training loss decreases steadily over optimization, and the evaluation loss reaches its minimum of 0.0387 at step 210. The overall trajectory indicates stable convergence during fine-tuning.}
    \label{fig:loss_curve}
\end{figure}

The training dynamics are correspondingly stable. \cref{fig:loss_curve} shows that both training and evaluation loss decrease smoothly throughout optimization, with the final train loss reaching 0.0868 and the best evaluation loss reaching 0.0387 at step 210. The evaluation curve tracks the training trend without late-stage instability, which suggests that the fine-tuning procedure is well behaved on this task. Taken together, these results indicate that \sysName can support not only evaluation of agent safety failures, but also the development of downstream guard models that are substantially better aligned with the threat patterns present in realistic and diverse agent settings.

\begin{table}[t]
\centering
\caption{Guard-model performance on R-Judge.}
\label{tab:guard_rjudge}
\scriptsize
\setlength{\tabcolsep}{10pt}
\begin{tabular}{lrrr}
\toprule
\textbf{Model} 
& \textbf{Acc} & \textbf{Rec.} & \textbf{F1} \\
\midrule
LlamaGuard3            & 53.7 & 100.0 & 69.5 \\
NemoGuard              & 54.4 & 40.6  & 48.5 \\
Qwen3-Guard            & 59.4 & 32.3   & 45.8  \\
BraveGuard             & 57.8 & 91.2  & 69.7 \\
\textbf{Finetuned (Ours)} & \textbf{61.7} & 77.9 & 68.4 \\
\bottomrule
\end{tabular}
\end{table}

To examine whether the safety judgments learned from \sysName transfer beyond our benchmark, we further evaluate the fine-tuned guard model on R-Judge\cite{yuan2024r}, a separate safety-classification benchmark with a different distribution of prompts and decision boundaries. The results are reported in \cref{tab:guard_rjudge}. Several observations are notable. First, the fine-tuned model achieves the highest accuracy at 61.7\%, exceeding all off-the-shelf baselines, which suggests that training on \sysName does not merely overfit to the annotation conventions or execution structure of our own benchmark. Instead, it appears to improve the model's ability to separate harmful from non-harmful cases under distribution shift. Second, the recall of the fine-tuned model reaches 77.9\%, which is lower than the extremely aggressive settings of LlamaGuard3 and BraveGuard\cite{feng2026braveguard} but substantially higher than NemoGuard and Qwen3-Guard. This places our model at a more balanced operating point between sensitivity and overprediction. Third, although BraveGuard attains the highest F1 score by a small margin, its lower accuracy suggests a stronger tendency to label examples as unsafe. By contrast, our fine-tuned model delivers the most accurate overall judgments while maintaining competitive F1, indicating that \sysName-derived supervision yields a detector with improved calibration rather than simply higher positive prediction rates. Taken together, these results provide preliminary evidence that the safety signals captured by \sysName are not purely benchmark-specific and can support guard models that generalize to external evaluation settings.

\section{Conclusion}
\label{sec:conclusion}

We introduced \sysName, a framework that operationalizes agent safety testing through executable safety cases, adaptive runtime interaction, and evidence-grounded verification over environment state. Experiments on four production agent frameworks reveal substantial safety weaknesses across diverse risks, attack methods, and environments, while guard models fine-tuned on \sysName-Bench generalize more effectively than strong off-the-shelf baselines. As agents become more autonomous and tool-integrated, safety evaluation must evolve from static benchmarking toward modular, executable testing infrastructure; we hope \sysName helps establish such infrastructure for future agent assurance.

\newpage
\bibliographystyle{IEEEtran}
\bibliography{main}

\vspace{12pt}
\color{red}

\end{document}